\documentclass[letterpaper, 10 pt, conference]{ieeeconf}  

\IEEEoverridecommandlockouts                              

\usepackage{listings}
\usepackage{float}
\usepackage[normalem]{ulem}
\usepackage[utf8]{inputenc}
\usepackage{bbding}
\usepackage[T1]{fontenc}
\usepackage[textsize=scriptsize, disable]{todonotes} 
\usepackage{lipsum}
\usepackage{algorithm}
\usepackage{algpseudocode}
\usepackage{url}
\usepackage{blindtext}
\usepackage[separate-uncertainty = true]{siunitx}
\usepackage{color,soul}
\usepackage{svg}
\usepackage{booktabs}
\usepackage{amsmath} 
\usepackage{amsfonts}
\usepackage{amssymb}  
\usepackage[nolist]{acronym}
\usepackage{courier}
\usepackage{xcolor}

\newcommand{\insertref}[1]{\todo[color=green!40]{#1}}

\newcommand{\skiros}{SkiROS2}

\begin{acronym}
\acro{api}[API]{application programming interface}
\acro{bt}[BT]{behavior tree}
\acroplural{bt}[BTs]{behavior trees}
\acro{dt}[DT]{digital twin}
\acroplural{dt}[DTs]{digital twins}
\acro{ebt}[eBT]{extended behavior tree}
\acroplural{ebt}[eBTs]{extended behavior trees}
\acro{fsm}[FSM]{finite state machine}
\acroplural{fsm}[FSMs]{finite state machines}
\acro{gui}[GUI]{graphical user interface}
\acroplural{gui}[GUIs]{graphical user interfaces}
\acro{htn}[HTN]{hierarchical task network}
\acro{mes}[MES]{manufacturing execution system}
\acroplural{mes}[MES]{manufacturing execution systems}
\acro{mg}[MG]{motion generator}
\acroplural{mg}[MGs]{motion generators}
\acro{owl}[OWL]{Web Ontology Language}
\acro{pddl}[PDDL]{Planning Domain Definition Language}
\acro{rdf}[RDF]{Resource Description Framework}
\acro{rl}[RL]{reinforcement learning}
\acro{ros}[ROS]{Robot Operating System}
\acro{wm}[WM]{world model}
\end{acronym}

\newfloat{lstfloat}{tbp}{lop}
\floatname{lstfloat}{Listing}

\definecolor{lightgray}{rgb}{0.98,0.98,0.98}
\definecolor{commentgreen}{RGB}{2,112,10}
\definecolor{eminence}{RGB}{108,48,130}
\definecolor{weborange}{RGB}{255,165,0}
\definecolor{frenchplum}{RGB}{129,20,83}
\definecolor{ceruleanblue}{rgb}{0.16, 0.32, 0.75}

\title{\LARGE \bf
SkiROS2: A skill-based Robot Control Platform for ROS
}

\author{Matthias Mayr$^{1}$, Francesco Rovida$^{2}$ and Volker Krueger$^{1}$
\thanks{This work was partially supported by the Wallenberg AI, Autonomous Systems and Software Program (WASP) funded by Knut and Alice Wallenberg Foundation.
}
\thanks{$^{1}$Department of Computer Science, Faculty of Engineering (LTH), Lund University, SE~221~00 Lund, Sweden. E-mail: <firstname>.<lastname>@cs.lth.se.}
\thanks{$^{2}$ RiACT ApS, A.C. Meyers Vaenge 15, 2450, Copenhagen, Denmark. Email: f.rovida@riact.eu
	}%
 }

\begin{document}
\maketitle
\thispagestyle{empty}
\pagestyle{empty}

\begin{abstract}
The need for autonomous robot systems in both the service and the industrial domain is larger than ever. In the latter, the transition to small batches or even “batch size 1” in production created a need for robot control system architectures that can provide the required flexibility. Such architectures must not only have a sufficient knowledge integration framework. It must also support autonomous mission execution and allow for interchangeability and interoperability between different tasks and robot systems.
We introduce \textit{SkiROS2}, a skill-based robot control platform on top of ROS. SkiROS2 proposes a layered, hybrid control structure for automated task planning, and reactive execution, supported by a knowledge base for reasoning about the world state and entities.
The scheduling formulation builds on the extended behavior tree model that merges task-level planning and execution. This allows for a high degree of modularity and a fast reaction to changes in the environment. The skill formulation based on \mbox{pre-,} hold- and post-conditions allows to organize robot programs and to compose diverse skills reaching from perception to low-level control and the incorporation of external tools.
We relate \skiros~to the field and outline three example use cases that cover task planning, reasoning, multisensory input, integration in a manufacturing execution system and reinforcement learning.

\end{abstract}

\section{Introduction}
\label{sec:introduction}
Modern intelligent robots require an increasing system complexity in order to perform the increasingly complex tasks demanded from them. Especially in view of greater autonomy, this complexity needs to be matched by the system architecture used to program and control the robots. With more and more integrated systems that coordinate different partial solutions, there is a need for interoperability and a common framework for communication, control and task planning.

In the industrial robotics field this can be seen in the Industry 4.0 movement that advocates such a transition, but several aspects of current practice present barriers to it. Many robot control systems currently rely heavily on "implicit" knowledge representation. Typically this is implemented by using if-else statements in the actual code. This often inhibits the growth of a control platform, as well as the interchangeability and interoperability between different tasks or robot systems, as this knowledge is hidden and often only known to the programmer herself/himself. Furthermore, vendor lock-in into the robot programming platform of a specific manufacturer is widespread.

These barriers have also been identified in the past and early platforms such as \textit{ClaraTy}~\cite{volpe2001claraty} or \textit{LAAS}~\cite{bensalem2009designing} provided first architectures. For knowledge integration frameworks, the system around the \textit{Rosetta} ontology~\cite{stenmark13faiaa, stenmark15racm} and \textit{Knowrob}~\cite{tenorth092iicirs, tenorth13tijorr} created a strong foundation. However, the former is not publicly available and the latter targets the different needs of service robotics.


In this context, we introduce \textit{SkiROS2}, a skill-based robot control framework for ROS. SkiROS2 is open source and the successor platform of SkiROS1~\cite{rovida2017skiros}. It utilizes knowledge representation in a \ac{rdf} graph that supports an open and explicit formulation of knowledge. The skill model is based on \textit{pre-conditions} that are checked before a skill is executed, \textit{hold-conditions} that also need to be satisfied while the skill is running, and \textit{post-conditions} that are checked after the skill execution. SkiROS2 supports reasoning to infer skill parameters, and it allows the implementation and integration of custom reasoners, such as a spatial reasoner. Built-in task planning allows it to utilize robot capabilities to automatically construct a planning domain and problem description without manual input from a domain expert. Furthermore, it is capable of orchestrating multiple robot systems.

\begin{figure}[tpb]
	{
		\setlength{\fboxrule}{0pt}
		\framebox{\parbox{3in}{
				\centering
				\includegraphics[width=1.0\columnwidth]{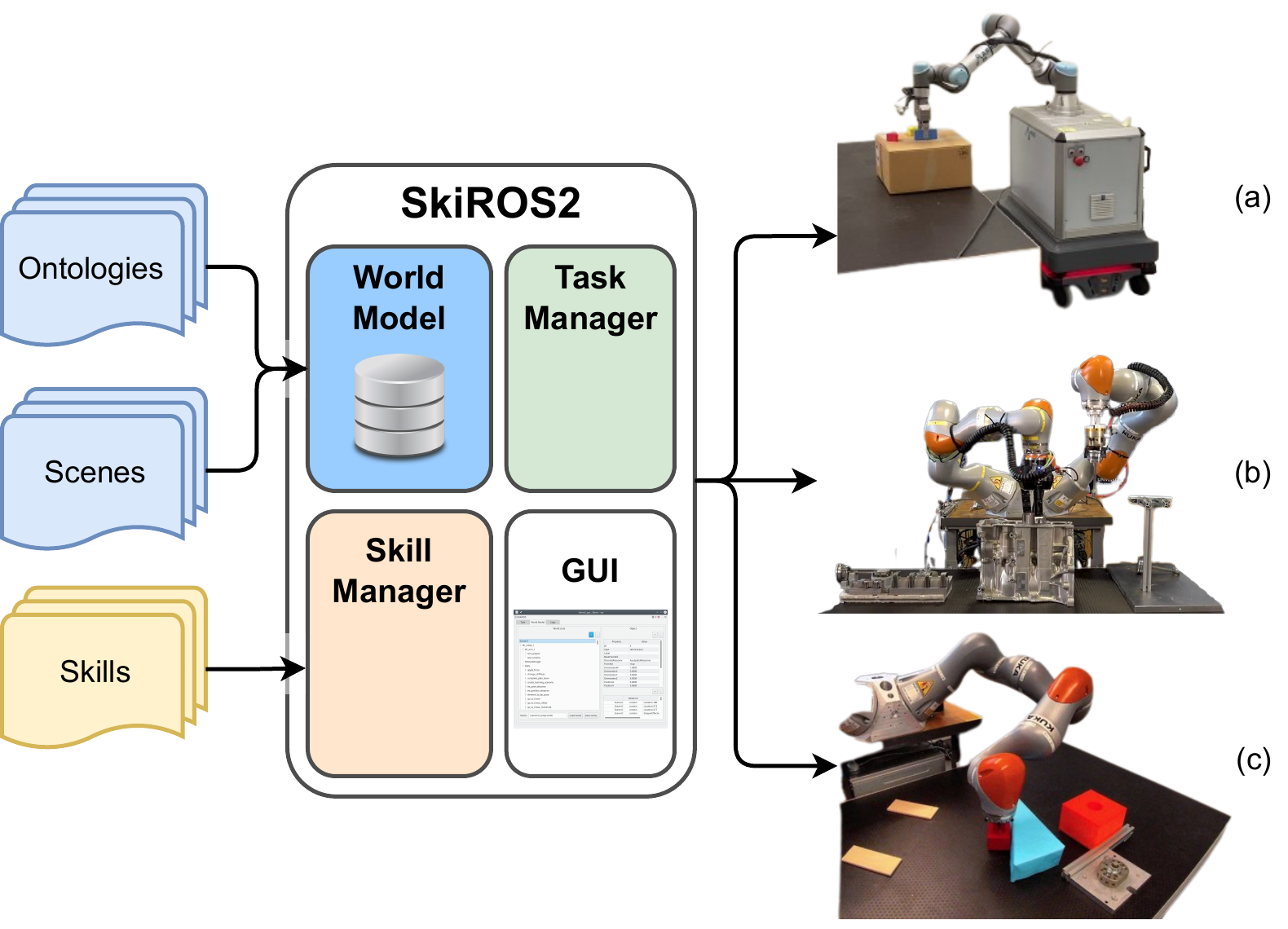}}
		}
	}
	\caption{\skiros~can be used with different ontologies, scene descriptions and skills to solve a variety of tasks with different robot platforms. (a):~Task-level planning with the mobile manipulation of objects. (b): Precise and sensitive piston insertion. (c): Learning to push by combining planning, reasoning and reinforcement learning.}
	\label{fig:overview}
\end{figure}

In this paper, we discuss the requirements for our system architecture and assess the state-of-the-art in the field and we show how our solution \skiros~fulfills these requirements. We introduce its individual modules such as the skill manager, the task manager and the world model. 

We think that this approach establishes a modern system architecture for intelligent autonomous robot systems that allows to design and organize robot skills for advanced tasks.

\section{Background and Related Work}
\label{sec:related-work}
This section introduces related work on the axes of robot control strategies, cognitive approaches as well as knowledge integration. Finally, a brief comparison with other platforms is provided.

\subsection{Robot Control Strategies}
\label{sec:robot-control}
Three control strategies dominate the research community: deliberative, reactive, and hybrid control~\cite{kortenkamp2016robotic}.

\subsubsection{Deliberative Systems}
\insertref{insert references}
This architecture follows the "sense, plan, act" paradigm. An observation of the world is translated into a symbolic representation~\cite{kortenkamp2016robotic}. Reasoning is then utilized to plan a sequence of actions. While the actions are executed new observations are not taken into account. This typically leads to slower reaction times, since the sequence of sensing and planning can often be time-consuming.

\subsubsection{Reactive Systems}
Reactive systems utilize a concurrently running finite state machines (FSM)\insertref{reactive systems}. These FSMs directly connect the input to the output. An important distinction is that they do not build up an internal representation. Their extension, behavior-based systems, also allows to assign priorities to behaviors and therefore inhibit output of low-priority ones.

\subsubsection{Layered Hybrid Systems}\insertref{insert references}
In order to utilize the advantages of both aforementioned systems, reactivity and planning, layered hybrid systems were formulated. Typically, a higher planning level determines a mid- or long-term strategy, and a lower level is able to react directly to observations. Most often a synchronization layer is added to bridge between high-level reasoning and low-level control. Most modern robot control systems follow this approach.

\subsection{Cognitive Systems}
\label{sec:cognitive-systems}
Besides having an architecture that allows tasks to be performed, another important question is what cognitive abilities are necessary and how these abilities are structured. Cognition can be defined as the process of a system to perceive the environment, act to pursue goals, anticipate the outcome of events, adapt the actions to changing circumstances and learn from experiences.

The \textit{cognitivist approach} performs processing to obtain a symbolic knowledge representation~\cite{kortenkamp2016robotic, vernon2011roadmap}. This abstracted symbolic representation of the world allows to reason about it. The representation is typically designed and interpreted by humans, which means that it can be understood and complemented with knowledge from other sources.

In contrast to that, the \textit{emergent paradigm} means an inductive organization from observations~\cite{vernon2011roadmap}. Machine learning techniques and the automatic induction of ontologies~\insertref{ontology introdcution?} are part of this field. It comes at the cost of being dependent on experiences.

\textit{Hybrid approaches} aim to combine both approaches, explicit knowledge representation and learning from experiences~\cite{vernon2011roadmap}.

For our work, the cognitivist approach is the most relevant. However, a recent line of research extends it to a hybrid approach by incorporating learning from experiences into the system~\cite{mayr2022combining, mayr22skireil, ahmad2022generalizing}.

\subsection{Knowledge Integration}
\label{sec:knowledge-integration}
An important aspect of autonomous systems is the type of knowledge integration. It decides on the support for reasoning and inference, the possibility of integrating encyclopedic knowledge, and the expressiveness. The latter should be weighted against the efficiency of the reasoning algorithms~\cite{kortenkamp2016robotic}.

The available formalisms in the field of robotics include model logic, temporal logic, and predicate logic such as Prolog and Golog. Description logics are widely used to define ontologies. In particular the \ac{owl} that is based on the \acf{rdf} gained a lot of attention. A recent review and comparison of ontology-based approaches is presented in~\cite{olivares2019review}. Projects such as \textit{Knowrob}~\cite{tenorth092iicirs, tenorth13tijorr} created a strong foundation in the field of service robotics, while, for example, \textit{Rosetta}~\cite{stenmark13faiaa} is explicitly designed for the industrial domain.

\subsection{Robot Control Platforms}
We relate our work to other frameworks that are available as open-source software or provide clear guidelines for their task planning, task execution, and knowledge integration. Table~\ref{tab:comparison} presents them along with criteria such as the application area, knowledge modeling and task planning. Most of the related frameworks explicitly target the industrial domain~\cite{stenmark13faiaa, stenmark15racm, paxton2017costar, huckaby2014knowledge} and~\cite{balakirsky2013knowledge}. Others, such as \textit{Knowrob + CRAM}~\cite{tenorth092iicirs, tenorth13tijorr} or \textit{CAST}~\cite{wyatt2010self} address the different needs of service robotics instead. Task-level planning is implemented by most of the frameworks. COSTAR~\cite{paxton2017costar} implements visual neural network-based planning~\cite{paxton2019visual} while~\cite{schillinger2016human} offers an interface that can be used by planners. For the scheduling we see that only few implement reactive methods such as \acp{bt}. As a middleware, most frameworks use the \acf{ros}, which is advantageous for creating interoperability and using network effects. Many, but not all platforms are available as open-source software for other researchers to study, use, and improve.
The comparison with existing frameworks shows that \skiros~has a unique combination attributes that enables it to be a modern autonomous robot control platform.

\begin{table*}[]
	\caption{A comparison of different robot control platforms. The existence or lack of a feature is shown with \Checkmark and \XSolidBrush. The "-" indicates a lack of information.}
	\begin{tabular}{@{}llllllll@{}}
		\toprule
		\textbf{Project Name/Group} & \textbf{Application} & \textbf{Knowledge Modeling} & \textbf{Task Planning} & \textbf{Scheduling} & \textbf{Middleware} & \textbf{Open Source} \\ \midrule
		SkiROS2                      & Industrial           & OWL-DL                      & PDDL                   & eBT                 & ROS                 & \Checkmark                \\
		COSTAR~\cite{paxton2017costar}                      & Industrial           & -                           & Visual                   & BT                  & ROS                 & \Checkmark            \\
		GTax~\cite{huckaby2014knowledge}                        & Industrial           & SysMl                       & PDDL                   & SysMl               & -                   & \XSolidBrush                 \\
		Balakirsky et al.~\cite{balakirsky2013knowledge}           & Industrial           & OWL-DL                      & PDDL                   & CRCL                & ROS                 & \XSolidBrush                 \\
		Stenmark et al.~\cite{stenmark13faiaa, stenmark15racm}            & Industrial           & OWL-DL                      & PDDL                   & State charts        & -                   & \XSolidBrush                 \\
		Knowrob/CRAM/EASE~\cite{tenorth092iicirs, tenorth13tijorr}          & Service              & Prolog, OWL-DL              & CPL                    & CPL                 & ROS                 & \Checkmark                \\
		CAST~\cite{wyatt2010self}                        & Service              & Proxies                     & MAPL                   & MAPL                & BALT               & partially            \\
		LAAS~\cite{bensalem2009designing}                         & General              & -                           & lxTeT                  & Open-PRS            & Bip/GenoM           & partially            \\
		ClaraTy~\cite{volpe2001claraty}                     & General              & -                           & Corba                  & TDL                 & -                   & \Checkmark                \\
		SmartMDSD~\cite{stampfer2016smartmdsd}                   & General              & -                           & SmartTdl               & SmartTdl            & SmartSoft           & \Checkmark                \\ 
		FlexBE~\cite{schillinger2016human}          & General & \XSolidBrush & (Synthesis Interface) & FSM & ROS & \Checkmark \\
		\bottomrule
	\end{tabular}
	\label{tab:comparison}
\end{table*}

\subsection{Behavior Trees}
\label{sec:bt}
A \acf{bt} is a formalism for the representation and execution of procedures. \acp{bt} have emerged in the gaming industry, but are also becoming widespread in robotics~\cite{iovino2022survey}.
A \ac{bt} is a directec acyclic and rooted graph consisting of nodes and edges~\cite{colledanchise17bt}. The root node is used to periodically inject an enabling signal called \textit{tick} that traverses through the tree according to the conditions, state of skills and their connectors called \textit{processors}. These processors allow to link child nodes in different procedural ways. Examples are a sequence (logical AND) or a selector (logical OR). When the \textit{tick} reaches a leaf node, it executes one cycle of the action or condition. Actions can modify the system configuration and return one of the three signals \textit{success}, \textit{running} or \textit{failure}. Condition checks are atomic and can only return \textit{success} or \textit{failure}. To pass information between different nodes, a common approach is to use a set of shared variables on a \textit{blackboard}. For a full formalization of \acp{bt} in the context of robotics, we refer to~\cite{colledanchise17bt}.

The classical \ac{bt} formulation is complemented by a formalism to define \textit{\acp{ebt}} in~\cite{rovida2017extended}. In \acp{ebt}, scripted and planned procedures are merged into a unified representation, so that an \ac{ebt} describes both the execution and its effects on the world state. This is achieved by combining the flexibility of \acp{bt} with \ac{htn} planning. In contrast to classical \ac{bt}, the pre- and post-condition nodes are embedded into the \ac{ebt} to use them for task-level planning. To achieve real modularity, \ac{ebt} allows procedural abstraction by allowing different implementations for the same type of action. Additional pre-conditions then allow to choose the right implementation to use at run-time. For example, select a different opening strategy depending if you need to open a manual door or an automatic one that opens with a button.  Furthermore, the \ac{ebt} formalism allows to optimize the execution of the planned sequence~\cite{rovida2017extended}.

\section{Design Considerations}
\label{sec:design-considerations}
On the basis of common applications in the field and our own experience, we identified the relevant requirements and design considerations. This section presents and discusses them in the context of robot control systems.

\subsection{Control Strategy}
Out of the main branches of robot control structures introduced in Section~\ref{sec:robot-control}, the layered hybrid approach is the most applicable to the target domain: longer-term task goals require planning and reasoning while the execution should be able to react quickly to observations from the real world. Therefore, \skiros~uses a deliberate planning level and a lower-level implementation with \ac{bt}~\cite{colledanchise17bt}.

\subsection{Multi-Robot Orchestration}
Challenging modern Industry 4.0 tasks are rarely content with a single robot.
In addition, because of the importance of collaboration with humans, a robot control platform needs to be able to orchestrate several actuators simultaneously. In a multi-robot setup it is important that all actuators maintain a coherent world state to prevent failures and synchronization problems. \skiros~allows to start an arbitrary number skill managers for different robot systems that can share a single \ac{wm}, thus being able to simultaneously orchestrate a fleet of robots that have a common understanding of the world.

\subsection{Knowledge Representation}
To avoid an implicit representation of knowledge, for example in the form of "if, else" statements in code, it is important to have means to \textit{explicitly} formulate and organize knowledge. Autonomous robot systems can also benefit from knowledge integration for abstract reasoning and especially in the industrial robotics domain, structured knowledge is often available\insertref{knowledge in industry}.
There are many logic formalisms\insertref{logic formalisms} and the choice needs to balance \textit{expressivity} and \textit{efficiency}. Furthermore, we find the usability and availability of ontologies, the set of concepts and their relation in a target domain, to be important.
Therefore, \skiros~uses the \ac{owl} and a set of established ontologies such as \textit{Core Ontology for Robotics and Automation (CORA)}.

\subsection{Manufacturing Execution Integration}
A platform for industrial robot skill execution needs to be able to interact with higher-level systems such as a \acf{mes}. Although the manual or automated start of individual skills is important, the concept of a skill-based platform really excels when higher-level goals can be sent to robot control systems~\cite{krueger16ieee}. A robot control system then needs to plan and execute autonomously. Furthermore, the execution state and modification of a shared understanding of the world such as a \ac{wm} or a \acf{dt} must be reported back~\cite{krueger16ieee}.

\subsection{Stakeholders}
Robot control systems have different stakeholders that need to be addressed. Besides special needs for a vertical integration described in the previous section, users also have different expectations and needs. Most systems differentiate between developers and end users. Developers are assumed to have an in-depth understanding of the matter and need tools to support their processes. With a robot control platform, it can be assumed that developers design and implement skills. They can also understand and form ontologies.
On the other hand, end users need lower entry hurdles. In the robotics context, this usually means addressing them with a \acf{gui} that abstracts away the underlying processes.

\subsection{Middleware}
Finally, the middleware as a communication system and \ac{api} are an important choice for every larger software project. The robotics domain with its different actuators, sensors and software solutions is particularly diverse and support for a wide variety of robotic systems can easily surpass the capabilities of companies and research groups.

Although there are other robotic middleware systems such as \textit{OROCOS}~\cite{bruyninckx2001open}, the \acf{ros}~\cite{Quigley09ros} became a commonly used and increasingly popular middleware solution for robotic systems and their programs. The ROS libraries are a good basis for integrated robot systems since they provide a standard communication and programming interface. Therefore, based on \ac{ros}, a control system such as \skiros~gets immediate access to a wide range of tools, drivers and trained users.

\section{Architecture of SkiROS2}
\label{sec:architecture}
\begin{figure}[tpb]
	{
		\setlength{\fboxrule}{0pt}
		\framebox{\parbox{3in}{
        \centering
		\includegraphics[width=0.9\columnwidth]{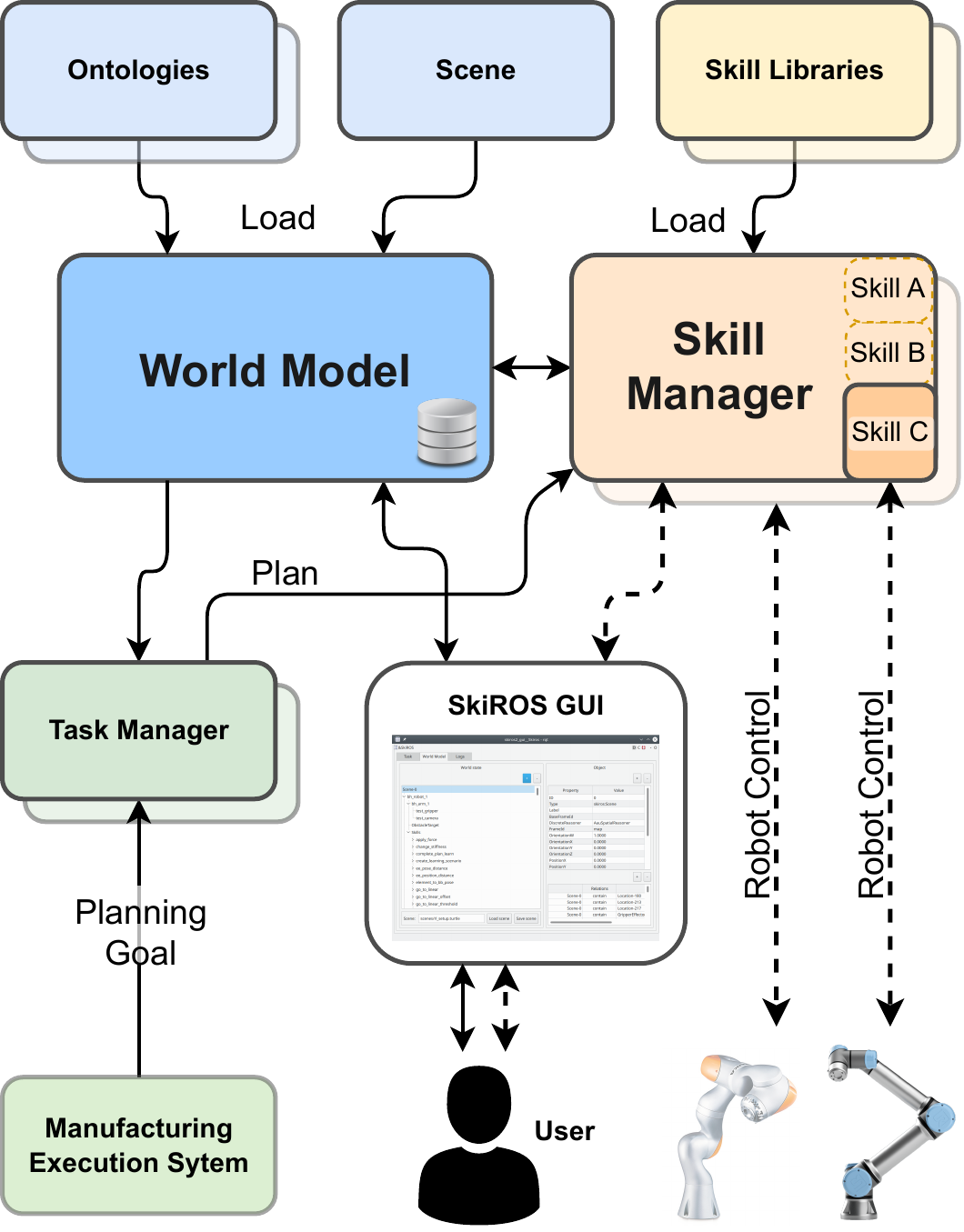}}
        }
	}
	\caption{An outline of the \skiros~architecture. The world model stores the knowledge about the relations, environment and the skills. The skill manager loads and executes the skills. Dashed lines show control flows and solid lines information flows. Shaded blocks indicate possible multiple instances.}
	\label{fig:architecture}
\end{figure}

This section outlines the architecture of the system shown in Fig.~\ref{fig:architecture}. The \textit{skill manager} and the \textit{\acf{wm}} form the core. One or more task managers can be used to accept high-level goals. During the prototyping phase, a \ac{gui} supports the users.

\subsection{Skill Model}
\label{sec:skills}
\begin{figure}[tpb]
	{
		\setlength{\fboxrule}{0pt}
		\framebox{\parbox{3in}{
				\centering
			\includegraphics[width=1.0\columnwidth]{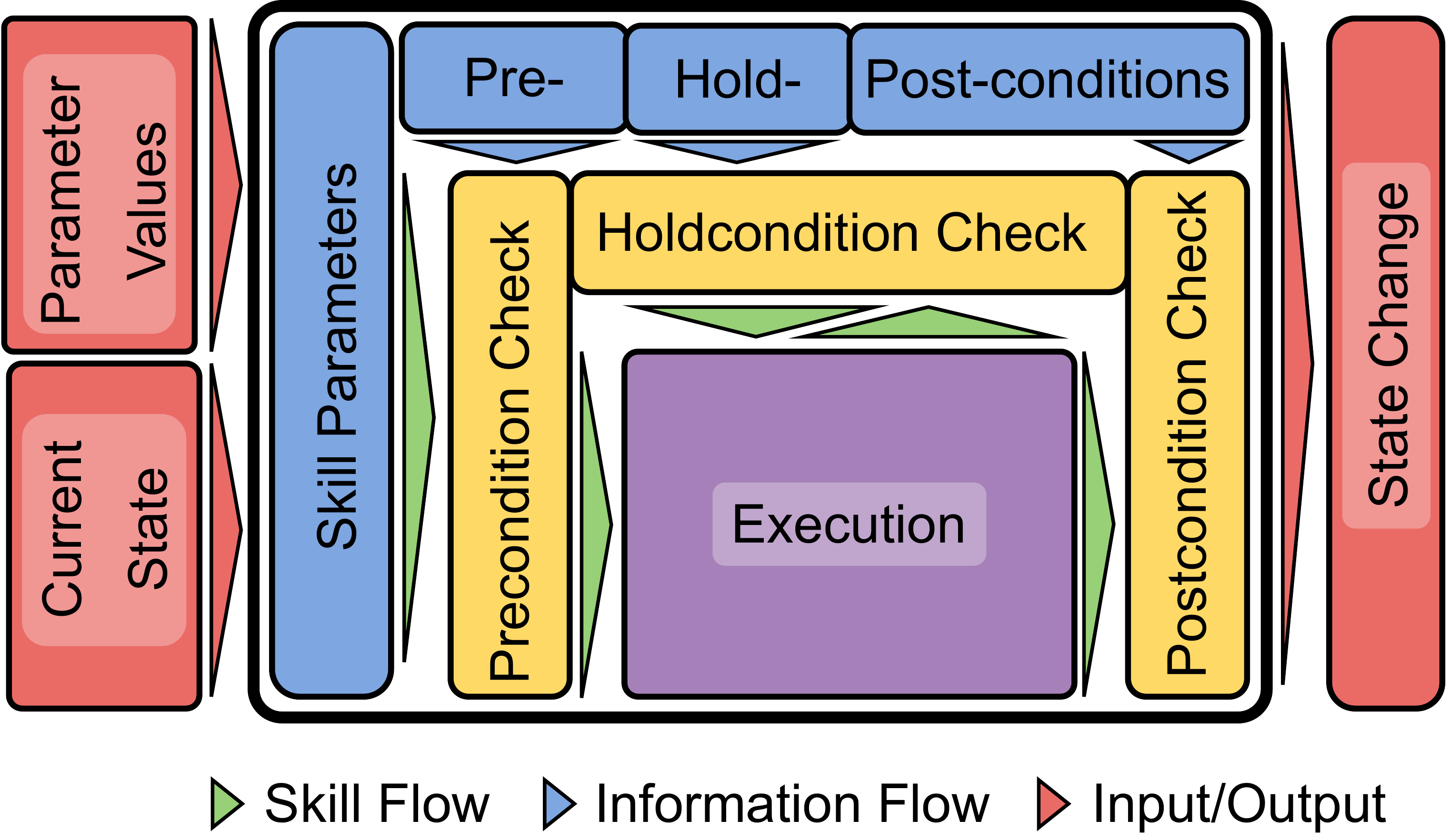}
		}}
	}
	\caption{The conceptual model of a skill in \skiros. \mbox{Pre-,} and hold-conditions ensure that the skill is only executed in the correct world state. Post-conditions check if the desired changes have been achieved.}
	\label{fig:skill}
\end{figure}

In this section we introduce the \skiros~skill model with its components and types of skills. We define a skill as a parametric procedure that changes the world from some initial state to some new state~\cite{bogh2012does}. This definition deliberately leaves room for a wide range of skills from deep learning-based object localization to lower-level motor control.
The skill execution flow is shown in Fig.~\ref{fig:skill}
In \skiros~skills can have different complexities: we refer here to primitive (atomic) skills and compound skills. The latter consist of any amount of primitive or compound skills. 

\textbf{Skill Description:} The skill description defines the actions of a skill on a semantic level. A skill description includes its zero or more parameters and zero or more pre-, hold- and post-conditions. An example is shown in Lst.~\ref{lst:pick-skill-short} Both primitive and compound skills always implemented exactly one skill description. However, an implementation is allowed to complement or further specify a skill description. This paradigm is useful to have multiple skill implementations for different hardware or specific scenarios. As an example for this, we can take a gripper actuation skill. A simple general skill description can be universal and have a boolean parameter "OpeningState" as well as a "Gripper" parameter that refers to a concept in the \ac{wm} such as \texttt{Element("rparts:GripperEffector")}. Different gripper hardware needs different implementations of this description. To enable this, \skiros~implements parametric polymorophism. A skill implementation for a specific gripper can then modify the skill description of the parameter "Gripper" to handle only a single type of gripper that would be a subtype of the concept "rparts:GripperEffector", such as "scalable:RobotiqGripper". When executing a compound skill that utilizes this gripper actuation skill description, \skiros~will automatically select the matching implementation.

\textbf{Skill Parameters:} The skill parameters define the input and output of a skill. Furthermore, they can also be used for reasoning together pre- and post-conditions. Parameters come in three flavors: 1) required, 2) optional, and \mbox{3) inferred}. In contrast to the optional parameters, the required parameters must be set to execute a skill. The inferred parameters do not need to be set, but can automatically be reasoned about with the pre- and post-conditions. As an example, the \textit{pick} skill in Lst.~\ref{lst:pick-skill-short} has a required parameter \textit{Arm}, could have an optional parameter \textit{arm maximum velocity} and has an inferred parameter \textit{Gripper}. If the parameter \textit{Arm} is known, we can utilize the relations in the \ac{wm} to automatically know which gripper is attached to it. The optional parameter for the maximum arm velocity can be used to overwrite a default value.

Skill parameters have either a fundamental data type such as \texttt{string}, \texttt{int}, \texttt{float}, \texttt{list}, \texttt{dict} or are \ac{wm} element in the form \texttt{Element("schema")}. The \textit{schema} refers to a concept that is defined in the ontologies, and during execution it is grounded to a specific instance of that concept. This way its associated attributes and relations can be utilized by the skills. In the \textit{pick} example above, the parameters \textit{gripper} and \textit{arm} would be \ac{wm} element types \texttt{Element("rparts:GripperEffector")} and \texttt{Element("scalable:Arm")}.

\textbf{Pre-, Hold- and Post-conditions:} 
Preconditions describe the necessary world state under which a skill is to be executed. Furthermore, they can be used to infer parameters of a skill such as the parameters \textit{Container} and \textit{Arm} in the example in Lst.~\ref{lst:pick-skill-short}. Preconditions are complemented by \textit{hold-conditions} that need to hold during the entire execution and are checked at every tick. Finally, the post-conditions specify the expected effects of a skill if the execution succeeded. Therefore, they are checked after the execution of the nodes has succeeded, and the skill will fail if the post-conditions do not hold.

All types of conditions have one of four types: 1) verification of relations in the \ac{wm} such as "Arm has a Gripper"; 2) check the existence of an element property, e.g. the gripper has an attribute that specifies the finger length; 3) compare the value of a property, for example having gripper finger longer than \SI{0.1}{\meter}; or finally 4) check for abstract relations in the ontology to verify that two elements are allowed to have a specified relation according to the ontology. In contrast to SkiROS1~\cite{rovida2017skiros}, primitives can also have pre-, post- and hold-conditions, which increases robustness and allows to infer parameters.

\textbf{Primitive Skills:} Primitive skills, or short primitives, are the atomic actions in the \skiros~skill model. They typically implement behaviors that actively change the world, such as opening a gripper or moving a robot arm. Lst.~\ref{lst:primitive} shows the primitive code skeleton that offers Python functions for the skill initialization, startup, execution, preemption and cleanup
. Primitives always have a skill description such as the one in Lst.~\ref{lst:pick-skill-short} that defines them on a semantic level and specifies the input and output parameters. When a skill is running, an \texttt{execute} function of the skill is called whenever the primitive is ticked as part of a \ac{bt} and it must return one of the three signals \textit{success}, \textit{running} or \textit{failure}.

\begin{lstfloat}
    \begin{lstlisting}[language=Python, basicstyle=\scriptsize, label={lst:pick-skill-short}, caption={The parameters, pre- and postconditions of a simple pick skill that can be automatically used for task-level planning.\vspace{0.3cm}}, xleftmargin=2.0ex]
class Pick(SkillDescription):
  def createDescription(self):
    # ======= Parameters =========
    self.addParam("Container",&\hspace*{-0.3em}& Element("skiros:Location"),&\hspace*{-0.3em}& ParamTypes.Inferred)
    self.addParam("Gripper", Element("rparts:GripperEffector"), ParamTypes.Inferred)
    self.addParam("Object", Element("skiros:Product"), ParamTypes.Required)
    self.addParam("Arm", Element("rparts:ArmDevice"), ParamTypes.Required)
    # ======= PreConditions =========
    self.addPreCondition(self.getPropCond("EmptyHanded", "skiros:ContainerState", "Gripper", "=", "Empty", True))
    self.addPreCondition(self.getRelationCond("ObjectInContainer", "skiros:contain", "Container", "Object", True))
    # ======= HoldConditions =========
    self.addHoldCondition(self.getRelationCond("RobotAtLocation", "skiros:at", "Robot", "Container", True))
    # ======= PostConditions =========
    self.addPostCondition(self.getPropCond("EmptyHanded", "skiros:ContainerState", "Gripper", "=", "Empty", False))
    self.addPostCondition(self.getRelationCond("Holding", "skiros:contain", "Gripper", "Object", True))
    \end{lstlisting}    
\end{lstfloat}

\begin{lstfloat}
    \begin{lstlisting}[language=Python, basicstyle=\scriptsize, label={lst:primitive}, caption={The code skeleton for a primitive skill. The functions allow to define the initialization, execution and reaction to a preemption. In line 3 it is stated which skill description (such as the one in Lst.~\ref{lst:pick-skill-short}) the primitive implements.\vspace{0.3cm}}, xleftmargin=2.0ex]
class my_primitive(PrimitiveBase):
  def createDescription(self):
    self.setDescription(MyPrimitive(), self.__class__.__name__)
  def onInit(self):
    """ Called once when loading skills - it is not loaded on False """
    return True
  def onPreempt(self):
    """ Called when skill is requested to stop. """
    return True
  def onStart(self):
    """ Called just before 1st execute """
    return True
  def execute(self):
    """ Main execution function. Returns: self.fail, self.step or self.success """
    return self.success("Executed")
  def onEnd(self):
    """ Called just after last execute OR preemption """
    return True
    \end{lstlisting}    
\end{lstfloat}

\textbf{Compound Skills:} In contrast to primitive skills, compound skills allow one to connect arbitrary amounts of other compound and primitive skills to define more complex behaviors. An example is given in Lst.~\ref{lst:compound}. Compound skills are modeled with \acp{bt} and their processors such as "sequential" or "parallel" that are introduced in Sec.~\ref{sec:bt}. This enables developers to fully utilize the modularity and reactiveness of the \acp{bt} by reusing existing skills and to formulate reactions to changes in the world.

\begin{lstfloat}
    \begin{lstlisting}[language=Python, basicstyle=\scriptsize, label={lst:compound}, caption={An example for a mockup pick skill that implements the example description in Lst~\ref{lst:pick-skill-short}. In line 7 the processor of the \ac{bt} is set to be sequential. In line 9 the parameter "Duration" of the skill "wait" is set to a concrete value and in line 10 parameter remappings are done.\vspace{0.3cm}}, xleftmargin=2.0ex]
class pick_fake(SkillBase):
  def createDescription(self):
    self.setDescription(Pick(), self.__class__.__name__)

  def expand(self, skill):
    """ In this function the BT is defined """
    skill.setProcessor(Sequential())
    skill(
      self.skill("Wait", "wait", specify={"Duration": 1.0}),
      self.skill("WmMoveObject", "wm_move_object",
        remap={"StartLocation": "Container", "TargetLocation": "Gripper"}),
    )
    \end{lstlisting}    
\end{lstfloat}
When including other skills in a compound skill, it is possible to explicitly set the parameters of the included skill if they are of any of the fundamental types. It is also possible to remap the parameters that are \ac{wm} elements. For example, the "pick" skill in Lst.~\ref{lst:pick-skill-short} has the parameter "Container". In the example in Lst.~\ref{lst:compound}, the primitive to update the world model ("wm\_move\_object") expects an input parameter "StartLocation". It is possible to map the references for these parameters, so they can be used in both skills. This can also be used to resolve naming conflicts.

\subsection{Skill Manager}
\label{sec:skill-manager}
Every robot has its own skill manager.
The skill manager loads a set of specified skills from the skill libraries. It initializes them and complements the \ac{wm} with semantic information such as the representation of the skill parameters and the skill conditions. After the initialization phase, the skill manager offers services to start, stop and debug skills, as well as monitor their execution. Since the skill manager is also executing the skills and accepts skill plans from the task manager, it is a core component of the platform.
Whenever a skill or skill sequence is started, the skill manager creates a new task with a unique ID. Within a task, the parameters of the executed skills are shared on a blackboard. This allows to exchange information such as calculated poses or even camera images between the skills.

\subsection{World Model}
\label{sec:world-model}
The \acl{wm} server stores the ontologies and the instances. As such, it loads the relevant ontologies that specify the known concepts (schemas) at the start. Furthermore, it can load a specific scene that contains the instances, also called vocabulary of the concepts that are specified in the ontologies. A typical scene includes the semantic description of the specific robot system, objects in the environment or known locations. The scene is complemented by the skill manager with information about the available skills.\insertref{sth about rdf}

The information in the \ac{wm} can be utilized for reasoning and for the parameterization of skills. It also exposes an \ac{api} that is used inside of the skills to read from and write to the \ac{wm}. Furthermore, custom reasoners can be plugged in. As an example for such reasoners, \textit{skiros2\_std\_lib} implements a spatial reasoner that can transform coordinates and calculate Allen intervals algebra~\cite{rovida2018motion}.

\subsection{Task Manager}
\label{sec:task-manager}
In Section~\ref{sec:design-considerations} we outlined the importance of the vertical integration into systems that provide tasks such as \ac{mes}. \skiros~addresses this by providing a sophisticated task planning integration that builds on the \acf{pddl}~\cite{aeronautiques1998pddl}. It utilizes the \textit{temporal fast downward planner}\footnote{\url{http://gki.informatik.uni-freiburg.de/tools/tfd/}}~\cite{eyerich2009using} to guarantee the finding of an optimal skill sequence. In contrast to other solution, \skiros~has a fully automated generation of the problem and planning domain based on the current state of the \ac{wm}. This includes the vocabulary of the scenes like the available robot as well as all the skills with their pre- and post-conditions. This generation drastically simplifies usability, as it is not necessary to maintain a separate planning domain.

The resulting plan can be sent to the skill manager and can become an executed task. It is then automatically converted into an executable \ac{ebt}~\cite{rovida2017extended} and the skill manager will expand the branches of the \ac{ebt} automatically.

\subsection{ROS Integration and User Interface}
\label{sec:gui}
\skiros~is well integrated into \acf{ros}~\cite{Quigley09ros}. While \skiros~utilizes \ac{ros} as a middleware itself, it also features many convenient integrations that ease the use and implementation of skills. The \ac{wm} and its elements are fully integrated with the \ac{ros} transformation system \textit{tf}. As such, \ac{wm} elements can be linked to frames that exist outside \skiros~and elements can be published as coordinate frames. On the skill level the \textit{skiros2\_std\_lib} provides a skill primitive that can easily turn any \ac{ros} \textit{action} into a skill.

The \skiros~\acf{gui} offers a drastically lowered entry hurdle for non-experts. Like many other \ac{ros} \acp{gui}, it is written with \ac{ros}' \textit{rqt} and can be used alongside them. In the different tabs, users get an overview of the loaded skills. Skill parameters can be changed and skills can be started and stopped. Additionally, the \ac{wm} view in the \ac{gui} provides full access to the content of the \ac{wm}. The vocabulary including all the properties and relations can be inspected as well as fully modified. New relations and properties can be added through easy-to-use dialogues. An integration with the visualization tool \textit{Rviz} allows to modify the pose of a \ac{wm} element.

\section{Use-Cases}
\label{sec:use-cases}
In this section, we discuss a selection of different use cases in which \textit{SkiROS2} is used as a robot control system for partially-structured tasks.

\subsection{Pick-place with a Mobile Robot}

In the pick-and-place scenario, a mobile robot (e.g. \textit{heron} in Fig.~\ref{fig:overview}a) is tasked to place an item at a new location that can either be at the same or at another workstation. In this scenario it is required that the robot can drive between workstations, actuate the arm as well as the gripper, and use the camera to compute the exact pose of the object. The skills can switch the control mode of the arm between compliant control for contact-rich tasks and position control for planned trajectories. The used skills have full integration with planning and it is sufficient to specify a goal in \ac{pddl} such as \texttt{skiros:contain skiros:locationB skiros:objectA}. As described in Section~\ref{sec:task-manager}, triggering the planning automatically generates the domain description and problem instance. A simplified example skill sequence is:
\begin{verbatim}
drive(workstationA)
pick(objectA)
drive(workstationB)
place(locationB)
\end{verbatim}

Lst.~\ref{lst:pick-skill-short} shows a simplified description of a \textit{pick} skill. The specified preconditions can be used to infer parameters at planning time, but also at run-time. If this example skill is called manually with \textit{SkiROS2} \ac{gui}, it is sufficient to specify the object to pick and the arm to use, and the rest of the parameters are automatically inferred using the ~\ac{wm} (Section~\ref{sec:world-model}).

\subsection{Dual-arm Piston Insertion}
The task in this use case is to perform a tight insertion of a piston into a real engine block. This insertion requires dual-arm manipulation as well as specialized tools since an assistive ring must be held for insertion. The skills used in this task heavily utilize the \ac{wm}, but are at the same time able to learn parts of the procedure, such as lifting poses, from kinesthetic teaching. Additionally, the pose of objects, such as the engine block, can be estimated using the camera mounted on the robot arm~\cite{grossmann2019continuous}. The motion skills implement a combination of a motion generator with \acp{bt}~\cite{rovida2018motion} and have extensive checks on the robot state such as applied forces and torques. The extensive use of those conditions allows aborting the execution if they are exceeded. A short video is available here\footnote{\url{https://www.youtube.com/watch?v=sTM0ih6faMs}}.

While the skills for this use-case do not have the necessary pre- and post-conditions to use them for task planning, the concept of these conditions is used to regularly check if the system is in the desired state. Furthermore, these skills are written so that they are preemptable~\cite{wuthier2021productive}. As an example for a preemption procedure we can consider a state in which an object is in the gripper. Aborting the current action and switching to a different task requires to place the object at an appropriate location first.

\subsection{Reinforcement Learning of Industrial Robot Tasks}
\label{sec:use-case-rl}
In a third use-case, \textit{SkiROS2} is used to learn industrial tasks with reinforcement learning (RL)~\cite{mayr2022combining, mayr22skireil, mayr22priors, ahmad2022generalizing}. Such tasks include pushing an object on a table (shown in Fig.~\ref{fig:overview}c) or learning a peg-insertion strategy. The skills used in these scenarios are written by domain experts and can be utilized by the task planner. However, there is no automated reasoning module for some of the skills that can be utilized to fully parameterize them for the task at hand. In this learning approach, the skill parameters in a skill sequence that cannot be parameterized are automatically identified. Then additional information about these learnable parameters, such as their type and upper and lower limits, is obtained from the \ac{wm} and used to automatically create a learning scenario description. In the next step, the learning procedure starts either on a real robot system or in simulation, and the RL framework calculates the rewards depending on the performance. When leveraging on learning in simulation, up to thousands of executions can be run to identify a robust and well-performing set of parameters. Finally, the best configurations are presented to the operator and a final set can be selected for production~\cite{mayr2022combining, mayr22skireil}. If the operator has an educated guess for good parameter values or experiences from similar tasks, this learning approach can also incorporate this information as priors to accelerate learning and increase safety~\cite{mayr22priors}. With~\cite{ahmad2022generalizing}, an extension of~\cite{mayr2022combining, mayr22skireil} is proposed to learn behaviors for a variety of task variations and allow zero-shot execution even for unseen task configurations.

\section{Conclusions}
Modern autonomous robotic systems require solutions for the fundamental requirements: knowledge organization, control structuring, multi-robot orchestration and integration with external systems. We outlined the requirements and introduced the skill-based robot control platform \skiros~and its core components. Its unique combination of features makes it suitable for intelligent autonomous robot control.
To show this, we outlined three example use cases that cover a wide variety of relevant challenges in robotics: Integration of diverse robotic systems, task-level planning and integration of vision and learning from users. As part of these \skiros~has also been shown to allow the combination of deductive methods such as reasoning with inductive methods such as learning to improve the performance of executions by interacting with the environment.

The \textit{SkiROS2} platform is fully open source and comes with documentation, tutorials and examples. Currently it integrates with ROS 1, but a ROS 2 version is in development. The code is available at: \url{https://github.com/RVMI/skiros2}

\addtolength{\textheight}{-3cm}

\section*{Acknowledgement}
The authors thank Bjarne Grossmann, David Wuthier, Faseeh Ahmad, Simon Kristoffersson Lind and Momina Rizwan for their contributions.

This work was partially supported by the Wallenberg AI, Autonomous Systems and Software Program (WASP) funded by the Knut and Alice Wallenberg Foundation.\\
This project has received funding from the European Union’s Horizon 2020 research and innovation program under grant agreement no. 723658, Scalable4.0.

\bibliography{root}

\begin{thebibliography}{10}
\providecommand{\url}[1]{#1}
\csname url@samestyle\endcsname
\providecommand{\newblock}{\relax}
\providecommand{\bibinfo}[2]{#2}
\providecommand{\BIBentrySTDinterwordspacing}{\spaceskip=0pt\relax}
\providecommand{\BIBentryALTinterwordstretchfactor}{4}
\providecommand{\BIBentryALTinterwordspacing}{\spaceskip=\fontdimen2\font plus
\BIBentryALTinterwordstretchfactor\fontdimen3\font minus
  \fontdimen4\font\relax}
\providecommand{\BIBforeignlanguage}[2]{{%
\expandafter\ifx\csname l@#1\endcsname\relax
\typeout{** WARNING: IEEEtran.bst: No hyphenation pattern has been}%
\typeout{** loaded for the language `#1'. Using the pattern for}%
\typeout{** the default language instead.}%
\else
\language=\csname l@#1\endcsname
\fi
#2}}
\providecommand{\BIBdecl}{\relax}
\BIBdecl

\bibitem{volpe2001claraty}
R.~Volpe, I.~Nesnas, T.~Estlin, D.~Mutz, R.~Petras, and H.~Das, ``The claraty
  architecture for robotic autonomy,'' in \emph{2001 IEEE Aerospace Conference
  Proceedings (Cat. No. 01TH8542)}, vol.~1.\hskip 1em plus 0.5em minus
  0.4em\relax IEEE, 2001, pp. 1--121.

\bibitem{bensalem2009designing}
S.~Bensalem, M.~Gallien, F.~Ingrand, I.~Kahloul, and N.~Thanh-Hung, ``Designing
  autonomous robots,'' \emph{IEEE Robotics \& Automation Magazine}, vol.~16,
  no.~1, pp. 67--77, 2009.

\bibitem{stenmark13faiaa}
M.~Stenmark and J.~Malec, ``Knowledge-{{Based Industrial Robotics}},''
  \emph{Frontiers in Artificial Intelligence and Applications}, pp. 265--274,
  2013.

\bibitem{stenmark15racm}
------, ``Knowledge-based instruction of manipulation tasks for industrial
  robotics,'' \emph{Robotics and Computer-Integrated Manufacturing}, vol.~33,
  pp. 56--67, Jun. 2015.

\bibitem{tenorth092iicirs}
M.~Tenorth and M.~Beetz, ``{{KNOWROB}} \textemdash{} knowledge processing for
  autonomous personal robots,'' in \emph{2009 {{IEEE}}/{{RSJ International
  Conference}} on {{Intelligent Robots}} and {{Systems}}}, Oct. 2009, pp.
  4261--4266.

\bibitem{tenorth13tijorr}
------, ``{{KnowRob}}: {{A}} knowledge processing infrastructure for
  cognition-enabled robots,'' \emph{The International Journal of Robotics
  Research}, vol.~32, no.~5, pp. 566--590, Apr. 2013.

\bibitem{rovida2017skiros}
F.~Rovida, M.~Crosby, D.~Holz, A.~S. Polydoros, B.~Gro{\ss}mann, R.~P. Petrick,
  and V.~Kr{\"u}ger, ``Skiros—a skill-based robot control platform on top of
  ros,'' \emph{Robot Operating System (ROS) The Complete Reference (Volume 2)},
  pp. 121--160, 2017.

\bibitem{kortenkamp2016robotic}
D.~Kortenkamp, R.~Simmons, and D.~Brugali, ``Robotic systems architectures and
  programming,'' \emph{Springer handbook of robotics}, pp. 283--306, 2016.

\bibitem{vernon2011roadmap}
D.~Vernon, C.~Von~Hofsten, and L.~Fadiga, \emph{A roadmap for cognitive
  development in humanoid robots}.\hskip 1em plus 0.5em minus 0.4em\relax
  Springer Science \& Business Media, 2011, vol.~11.

\bibitem{mayr2022combining}
M.~Mayr, F.~Ahmad, K.~Chatzilygeroudis, L.~Nardi, and V.~Krueger, ``Combining
  planning, reasoning and reinforcement learning to solve industrial robot
  tasks,'' \emph{IROS 2022 Workshop on Workshop on Trends and Advances in
  Machine Learning and Automated Reasoning for Intelligent Robots and Systems},
  2022.

\bibitem{mayr22skireil}
------, ``Skill-based multi-objective reinforcement learning of industrial
  robot tasks with planning and knowledge integration,'' in \emph{2022 IEEE
  International Conference on Robotics and Biomimetics (ROBIO)}.\hskip 1em plus
  0.5em minus 0.4em\relax IEEE, 2022, pp. 1995--2002.

\bibitem{ahmad2022generalizing}
F.~Ahmad, M.~Mayr, E.~A. Topp, J.~Malec, and V.~Krueger, ``Generalizing
  behavior trees and motion-generator (btmg) policy representation for robotic
  tasks over scenario parameters,'' \emph{2022 IJCAI Planning and Reinforcement
  Learning Workshop}, 2022.

\bibitem{olivares2019review}
A.~Olivares-Alarcos, D.~Be{\ss}ler, A.~Khamis, P.~Goncalves, M.~K. Habib,
  J.~Bermejo-Alonso, M.~Barreto, M.~Diab, J.~Rosell, J.~Quintas \emph{et~al.},
  ``A review and comparison of ontology-based approaches to robot autonomy,''
  \emph{The Knowledge Engineering Review}, vol.~34, p. e29, 2019.

\bibitem{paxton2017costar}
C.~Paxton, A.~Hundt, F.~Jonathan, K.~Guerin, and G.~D. Hager, ``Costar:
  Instructing collaborative robots with behavior trees and vision,'' in
  \emph{2017 IEEE international conference on robotics and automation
  (ICRA)}.\hskip 1em plus 0.5em minus 0.4em\relax IEEE, 2017, pp. 564--571.

\bibitem{huckaby2014knowledge}
J.~O. Huckaby, ``Knowledge transfer in robot manipulation tasks,'' Ph.D.
  dissertation, Georgia Institute of Technology, 2014.

\bibitem{balakirsky2013knowledge}
S.~Balakirsky, Z.~Kootbally, T.~Kramer, A.~Pietromartire, C.~Schlenoff, and
  S.~Gupta, ``Knowledge driven robotics for kitting applications,''
  \emph{Robotics and Autonomous Systems}, vol.~61, no.~11, pp. 1205--1214,
  2013.

\bibitem{wyatt2010self}
J.~L. Wyatt, A.~Aydemir, M.~Brenner, M.~Hanheide, N.~Hawes, P.~Jensfelt,
  M.~Kristan, G.-J.~M. Kruijff, P.~Lison, A.~Pronobis \emph{et~al.},
  ``Self-understanding and self-extension: A systems and representational
  approach,'' \emph{IEEE Transactions on Autonomous Mental Development},
  vol.~2, no.~4, pp. 282--303, 2010.

\bibitem{paxton2019visual}
C.~Paxton, Y.~Barnoy, K.~Katyal, R.~Arora, and G.~D. Hager, ``Visual robot task
  planning,'' in \emph{2019 international conference on robotics and automation
  (ICRA)}.\hskip 1em plus 0.5em minus 0.4em\relax IEEE, 2019, pp. 8832--8838.

\bibitem{schillinger2016human}
P.~Schillinger, S.~Kohlbrecher, and O.~Von~Stryk, ``Human-robot collaborative
  high-level control with application to rescue robotics,'' in \emph{2016 IEEE
  International Conference on Robotics and Automation (ICRA)}.\hskip 1em plus
  0.5em minus 0.4em\relax IEEE, 2016, pp. 2796--2802.

\bibitem{stampfer2016smartmdsd}
D.~Stampfer, A.~Lotz, M.~Lutz, and C.~Schlegel, ``The smartmdsd toolchain: An
  integrated mdsd workflow and integrated development environment (ide) for
  robotics software,'' \emph{Journal of Software Engineering for Robotics
  (JOSER)}, vol.~7, no.~1, pp. 3--19, 2016.

\bibitem{iovino2022survey}
M.~Iovino, E.~Scukins, J.~Styrud, P.~{\"O}gren, and C.~Smith, ``A survey of
  behavior trees in robotics and ai,'' \emph{Robotics and Autonomous Systems},
  vol. 154, p. 104096, 2022.

\bibitem{colledanchise17bt}
M.~Colledanchise and P.~Ögren, \emph{Behavior {{Trees}} in {{Robotics}} and
  {{AI}}: {{An Introduction}}}.\hskip 1em plus 0.5em minus 0.4em\relax Chapman
  \& Hall/CRC Press, 2017.

\bibitem{rovida2017extended}
F.~Rovida, B.~Grossmann, and V.~Kr{\"u}ger, ``Extended behavior trees for quick
  definition of flexible robotic tasks,'' in \emph{2017 IEEE/RSJ International
  Conference on Intelligent Robots and Systems (IROS)}.\hskip 1em plus 0.5em
  minus 0.4em\relax IEEE, 2017, pp. 6793--6800.

\bibitem{krueger16ieee}
V.~Krueger, A.~Chazoule, M.~Crosby, A.~Lasnier, M.~R. Pedersen, F.~Rovida,
  L.~Nalpantidis, R.~Petrick, C.~Toscano, and G.~Veiga, ``A vertical and
  cyber–physical integration of cognitive robots in manufacturing,''
  \emph{Proceedings of the IEEE}, vol. 104, no.~5, pp. 1114--1127, 2016.

\bibitem{bruyninckx2001open}
H.~Bruyninckx, ``Open robot control software: the orocos project,'' in
  \emph{2001 IEEE International Conference on Robotics and Automation (ICRA)},
  vol.~3.\hskip 1em plus 0.5em minus 0.4em\relax IEEE, 2001, pp. 2523--2528.

\bibitem{Quigley09ros}
M.~Quigley, B.~Gerkey, K.~Conley, J.~Faust, T.~Foote, J.~Leibs, E.~Berger,
  R.~Wheeler, and A.~Ng, ``Ros: an open-source robot operating system,'' in
  \emph{2009 IEEE International Conference on Robotics and Automation (ICRA)
  Workshop on Open Source Robotics}, Kobe, Japan, May 2009.

\bibitem{bogh2012does}
S.~B{\o}gh, O.~S. Nielsen, M.~R. Pedersen, V.~Kr{\"u}ger, and O.~Madsen, ``Does
  your robot have skills?'' in \emph{Proceedings of the 43rd international
  symposium on robotics}.\hskip 1em plus 0.5em minus 0.4em\relax VDE Verlag
  GMBH, 2012.

\bibitem{rovida2018motion}
F.~Rovida, D.~Wuthier, B.~Grossmann, M.~Fumagalli, and V.~Kr{\"u}ger, ``Motion
  generators combined with behavior trees: A novel approach to skill
  modelling,'' in \emph{2018 IEEE/RSJ International Conference on Intelligent
  Robots and Systems (IROS)}.\hskip 1em plus 0.5em minus 0.4em\relax IEEE,
  2018, pp. 5964--5971.

\bibitem{aeronautiques1998pddl}
C.~Aeronautiques, A.~Howe, C.~Knoblock, I.~D. McDermott, A.~Ram, M.~Veloso,
  D.~Weld, D.~W. SRI, A.~Barrett, D.~Christianson \emph{et~al.}, ``Pddl the
  planning domain definition language,'' \emph{Technical Report, Tech. Rep.},
  1998.

\bibitem{eyerich2009using}
P.~Eyerich, R.~Mattm{\"u}ller, and G.~R{\"o}ger, ``Using the context-enhanced
  additive heuristic for temporal and numeric planning,'' in \emph{Nineteenth
  International Conference on Automated Planning and Scheduling}, 2009.

\bibitem{grossmann2019continuous}
B.~Grossmann, F.~Rovida, and V.~Kruger, ``Continuous close-range 3d object pose
  estimation,'' in \emph{2019 IEEE/RSJ International Conference on Intelligent
  Robots and Systems (IROS)}.\hskip 1em plus 0.5em minus 0.4em\relax IEEE,
  2019, pp. 2861--2867.

\bibitem{wuthier2021productive}
D.~Wuthier, F.~Rovida, M.~Fumagalli, and V.~Kr{\"u}ger, ``Productive
  multitasking for industrial robots,'' in \emph{2021 IEEE International
  Conference on Robotics and Automation (ICRA)}.\hskip 1em plus 0.5em minus
  0.4em\relax IEEE, 2021, pp. 12\,654--12\,661.

\bibitem{mayr22priors}
M.~Mayr, C.~Hvarfner, K.~Chatzilygeroudis, L.~Nardi, and V.~Krueger, ``Learning
  skill-based industrial robot tasks with user priors,'' in \emph{2022 IEEE
  18th International Conference on Automation Science and Engineering
  (CASE)}.\hskip 1em plus 0.5em minus 0.4em\relax IEEE, 2022, pp. 1485--1492.

\end{thebibliography}
\bibliographystyle{bib/IEEEtran}

\end{document}